# Spatial embedding promotes a specific form of modularity with low entropy and heterogeneous spectral dynamics

Cornelia Sheeran[1,2*], Andrew S. Ham[1,3*], Duncan, E. Astle[1,4],
Jascha Achterberg[1,5+], Danyal Akarca[1,6,7+]

1. MRC Cognition and Brain Sciences Unit, University of Cambridge, UK
2. Center for the Study of Complex Systems, University of Michigan, USA
3. Harvard Medical School, Harvard University, USA
4. Department of Psychiatry, University of Cambridge, UK
5. Department of Physiology, Anatomy and Genetics, University of Oxford, UK
6. Department of Electrical and Electronic Engineering, Imperial College London, UK
7. I-X, Imperial College London, UK

*Co-lead authors
+Co-lead senior authors
Corresponding author: Danyal Akarca (d.akarca@imperial.ac.uk)

## Abstract
Understanding how biological constraints shape neural computation is a central goal of computational neuroscience. Spatially embedded recurrent neural networks provide a promising avenue to study how modelled constraints shape the combined structural and functional organisation of networks over learning. Prior work has shown that spatially embedded systems like this can combine structure and function into single artificial models during learning. But it remains unclear precisely how – in general – structural constraints bound the range of attainable configurations. In this work, we show that it is possible to study these restrictions through entropic measures of the neural weights and eigenspectrum, across both rate and spiking neural networks. Spatial embedding, in contrast to baseline models, leads to networks with a highly specific low entropy modularity where connectivity is readily interpretable given the known spatial and communication constraints acting on them. Crucially, these networks also demonstrate systematically modulated spectral dynamics, revealing how they exploit heterogeneity in their function to overcome the constraints imposed on their structure. This work deepens our understanding of constrained learning in neural networks, across coding schemes and tasks, where solutions to simultaneous structural and functional objectives must be accomplished in tandem.

## Introduction
Biological neural circuits are shaped by many constraints that can influence both the structural and functional organisation of those circuits. These constraints are a necessary consequence of multiple factors such as the network's geometry[1,2], development[3,4] and energy budget[5,6]. Understanding the relative influence of these constraints, and disentangling their specific contributions to system organisation, is a core challenge in neuroscience. For example, when will neurons in the brain exhibit pure versus mixed-selective codes[7,8]? When is it useful for neurons to exhibit dramatically varied physiological properties across their populations[9,10]? How do the limits of communication efficiency, imposed by connectivity and dynamics[11,12], affect resource rational behaviour[13,14]?

A promising approach to study the effects of constraints is to train neural networks to simultaneously solve tasks while dealing with some structural and/or functional constraint(s). The goal here, in contrast of simulating directly on empirical connectomes[15], is to apply abstracted constraints that are thought to mimic those occurring in nature. We can then test systematically whether the resulting artificial networks are more consistent with concurrent empirical observations of biological systems[16]. For example, recent evidence suggests that imposing spatial and communication constraints, by spatially embedding recurrent neural networks, can profoundly influence numerous outcomes of networks. Crucially, the characteristics produced by imposing these constraints mirror numerous observed features in neuroscience that are otherwise considered disparate[17]. These include modular network topologies[18], mixed selectivity[19] and spatially organised patterns of dynamic neural activity within the neural network. However, this prompts a more fundamental question: how do constraints influence the space of attainable configurations though learning? For example, do spatial constraints – by virtue of limiting physical connectivity – intrinsically limit the capacity of neural networks to solve problems[20]? Do communication constraints lead to relatively rarer network topologies[11,21]? Do network constraints limit the potential dynamics available during learning[22]? How may this be different in typical neural networks compared to more biologically veridical spiking neural networks[23]?

The current work addresses this by taking advantage of recent work that allows us to train biologically realistic spiking neural networks at scale[23,24] and examine the attainable configurations of both rate and spiking neural networks constrained via spatial embedding[17]. Our findings demonstrate that neural networks can be readily interpreted in context of the *entropy* of their neural weights and their eigenspectrum, given the constraints imposed on them[25,26]. Specifically, independent of task or coding scheme, we find that spatial embedding promotes a *highly specific form of modularity* with low entropy and heterogeneous spectral dynamics. We suggest that these findings are best understood when considering networks as exploiting their available heterogeneity within their bounded constraints.

# Methods

We aimed to investigate the topological, entropic and dynamical consequences of spatial and communication constraints on recurrent neural networks, in both rate and spiking setups, and across tasks. We start with a description of the different groups of constrained networks we tested before outlining the various graph theoretic and information theoretic measures employed to characterise the resultant networks. All code is available at https://github.com/DanAkarca/entropy_RNNs which includes demo training code and replication code.

## Modelled communication and spatial constraints

Our work builds on variations of spatially embedded recurrent neural networks (seRNNs)[17]. seRNNs are RNNs in which neurons are embedded within a discrete three-dimensional Euclidean space that places constraints on their learning. This embedding involves modulating the regularisation term such that rather than only promoting general sparsity (e.g., L1 regularisation), the network penalises other aspects of its neural structure, such as its spatial distances (reflecting the spatial cost of wiring)[18,20] or communication[27,28]. We term this general additional term the $L_{constraint}$, which constrains the learning process according to how it is defined. We subsequently construct the general learning objective according to the following equation:

$$L_{total} = L_{task} + \gamma L_{constraint}$$

where $L_{task}$ refers to the computed task error (see below for modelled tasks, depending on the rate or spiking implementation) and $\gamma$ denotes the regularisation strength. The larger $\gamma$, the larger the imposed constraint on the total loss, $L_{total}$. We tested four groups of networks, each trained via a subtle variation to probe the role of each constraint on network outcomes (**Table 1**). Importantly, baseline (L1) networks do not consider any topological structure of the network when learning to solve the task and are only constrained to promote sparsity. The spatial and communicability constrained networks consider the spatial distance and topological distance between neurons, respectively (and together, in the instance of seRNNs) and are also constrained to promote sparsity.

| Network group | $L_{constraint}$ |
|---|---|

| Baseline (L1) | $\sum |W|$ |
|---|---|
| Baseline + Space + Communicability (seRNN) | $\sum |W \odot D \odot C|$ |
| Baseline + Space only* | $\sum |W \odot D|$ |
| Baseline + Communicability only* | $\sum |W \odot C|$ |

**Table 1. Tested constraints within the spatially embedded recurrent neural network framework.** Our baseline networks were L1, where there is only a general bias toward sparse (low weight) solutions. The spatially embedded recurrent neural network (seRNN) additionally learns to optimise both the spatial and communicability matrix of the network. The last two network groups, denoted with an asterisk*, separate the space and communicability terms, and are referred to in the Supplement Material.

Here, $W$ refers to the weight matrix of the single hidden layer, $D$ refers to the Euclidean distance matrix, and $C$ is the network communicability. All networks were trained with a single hidden layer, constituted of 100 neurons, where inputs were provided to, and outputs read from, all hidden neurons. The $D$ matrix was constructed as in prior work by assigning each neuron a location within a 3D box (evenly spaced) with dimensions 5 x 5 x 4, and taking the Euclidean distance between neurons. The communicability was calculated as follows[28]:

$$C = e^{S^{-\frac{1}{2}} W S^{-\frac{1}{2}}},$$

where $S$ refers to the diagonal strength matrix of $W$. The normalisation of the communicability here has the effect of dampening the influence of aberrant weights of the network. The network communicability is a dynamical measure of information broadcasting which encompasses the set of all possible walks between nodes on a graph[21,22], and has shown to be particularly effective at bridging between structural and functional findings in neuroscience[29], likely owing to the fact that communicability is a more veridical approximation of communication in neural systems relative to connectivity alone[4,21].

For each of the network groups, we trained 1000 and 100 networks of gradually increasing $\gamma$ (following initial calculations for reasonable $\gamma$ values) for rate and spiking networks respectively. We next outline the details of how these objectives were implemented with rate and spiking recurrent neural networks, and the corresponding tasks we trained each to perform.

## Rate RNN Implementation

All rate-based RNNs were trained on a simple one-choice inference task, as in prior work. In this task, networks were presented a goal destination in one of four possible locations for 20 time-steps, at the corners of a 3 x 3 grid. After 20 time-steps, the goal was removed, and a delay was enacted for 10 time-steps. After this time delay, the network had 20 further time-steps with two choice options presented, followed by the network requiring to decide. A correct decision was made if the network picked the choice option closest to the goal location that was previously provided. As a result, this task requires that the network both remember and integrate information over the course of the trial.

Input goal and choice options were one-hot encoded, with added 0.1σ Gaussian noise. We present rate-based RNNs that, after training, could achieve >90% accuracy on the task (note that random chance is 25%). All networks were optimised using Adam[30] over ten training epochs, with a cross entropy loss function for computing $L_{task}$. More information on this implementation can be found in prior work[17].

# Spiking RNN Implementation

Spiking neural networks aim to emulate the activity of biological neurons by simulating firing, recreating the brain's temporal event-based dynamics and asynchronous communication. Given the disparities between rate- and spike-based coding mechanisms, we wanted to establish to what extent the different coding mechanism alters the network's learned configurations under analogous constraints. We constructed our spike-based RNNs using *snnTorch*[31], which utilises base *PyTorch* functions for creating connections between layers of neurons but uses its own custom neural models for instantiating the functions and properties of individual neurons and their recurrent connections within each layer.

We built our spiking neural networks using leaky integrate-and-fire (LIF) neurons characterised by the following discrete-time update equation:

$$V_{t+1} = \beta V_t + W I_{in_{t+1}} - S_t V_{thr}$$

Here, $V_{t+1}$ corresponds to the membrane potential of the neuron at time $t+1$, which depends on its previous potential $V_t$ weighted by a decay term, $\beta$, that simulates passive signal decay and leakage across the membrane. The input current $I_{in}$ to each neuron is weighted by the synaptic weights, $W$, and added to the membrane potential. The spiking mechanism $S_t$ is 1 after an action potential has fired, activating the reset mechanism by subtracting from the membrane potential. Due to previous work showing that heterogeneity in neuronal time constants improves performance, we allowed the delay term $\beta$ to be learned during training. This decay term is given by the following:

$$e^{-\frac{\Delta t}{\tau}},$$

where $\Delta t$ is the simulation time step of the model, set as 0.5 ms with a mean membrane time constant set at 20 ms. Decay rates were initialized by sampling randomly from a gamma distribution and clipped after every update. The upper bound was set to 0.995, translating to a time constant of 100 ms. According to the NeuroElectro database[32], 99.5% of biological neurons have membrane time constants below this upper threshold. Time constants greater than 1 stimulate exponential, uncontrolled growth in current and membrane potential. Very small values near 0 are also problematic as they induce unstable firing. A lower threshold constrained the time constant to a minimum value of 3Δt, where $\Delta t$ is the time step duration. As in prior work, we set the distribution of time constants to be free to evolve over training[10].

We trained these spike-based RNNs on a challenging neuromorphic classification task called the Spiking Heidelberg Digits (SHD)[33] task, which consists of 1000 spoken digits ranging from zero to nine in the English and German languages, from 12 unique speakers. The audio waveforms have been converted into spike trains using an artificial model of the inner ear and parts of the ascending auditory pathway. The SHD dataset has 8,156 training and 2,264 test samples. The individual samples varied in length between 0.24s and 1.17s. Inputs were fed into the spike-based RNN at every time step in the form of spikes, into a variable input layer containing the same number of neurons as stimulus dimensions. These spikes were linearly transformed by the input layer before being passed to the hidden layer. A layer of non-spiking readout neurons, corresponding to the number of classes in each dataset, then made decisions on stimulus identity based on the neuron receiving the highest level of current from the hidden layer. The task objective was to classify the dataset correctly. For the readout layer of neurons, spiking behaviour was required to be inhibited to network's determine the decision. If an action potential firing occurred, membrane potentials would reset and undergo a refractory period, obscuring which neuron had actually received the most current from the previous layer. To implement this, the final layer's neurons were assigned impossibly high threshold values, preventing them from spiking and allowing input current to accumulate over time. This was implemented using a cross entropy maximum membrane potential loss for computing $L_{task}$.

All spike-based RNNs were trained using Adam[30] with a learning rate of $10^{-3}$ and surrogate gradient descent[23,24] with steepness $\rho = 100$, for 50 epochs. We present spike-based RNNs that, after training, could achieve >45% accuracy on the task (note that random chance is 5%). In **Figure 1**, we provide a schematic of the overall methodological approach taken.

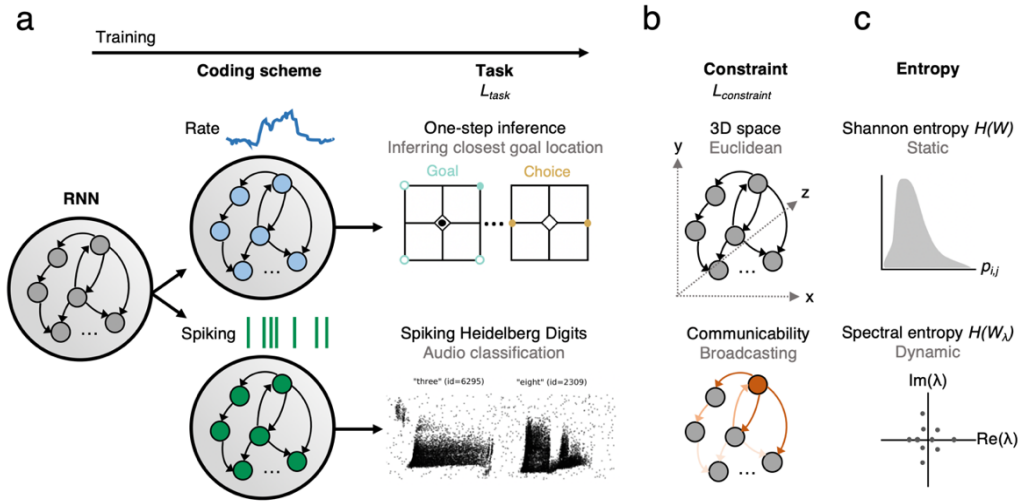

**Figure 1. Schematic of methodological approach**. **a** Recurrent neural networks (RNNs) were setup in two ways. The first setup consisted of a vanilla rate-based RNN, solving a one-step inference task as in prior work[17]. This task consists of observing a goal location on a grid (here, top right). Following a delay, the network is provided with two options, in which the network must decide which choice is closer to the goal location (here, right). The second setup consisted of a spike-based RNN, solving the SHD task. This task consists of a 20-class classification problem, with each problem consisting of spoken digits ranging from zero to nine in English or German. For both setups, the $L_{task}$ is the loss attributed to this task performance. **b** Two constraints were modelled in the RNNs, the first being 3D space (top) and the second being the network communicability which is a measure of broadcasting communication (bottom). **Table 1** outlines the complete learning objectives tested. We then tested various entropic measures of these resultant networks including the Shannon entropy of the weight matrix and entropic measures of the eigenspectrum.

## Network outcome measures

We next outline the measures used to evaluate network outcomes, of both rate and spiking neural networks, across each task and learning objective tested.

*Modularity*. The modularity quantifies the degree to which the network may be subdivided into such clearly delineated groups under an optimal community structure where nonoverlapping groups of nodes have a maximal the number of within-group edges and minimal number of between-group edges. Here, we quantified this using the maximum modularity Q statistic using the *modularity_dir* function in the Brain Connectivity Toolbox based on a deterministic algorithm at a default resolution parameter of 1[34,35].

*Shannon and Spectral entropy*. The entropy quantifies the amount of unpredictability or surprise in a set of possible outcomes. If a system is highly predictable (low uncertainty) it means the entropy will be low and will exhibit a degree of clustering around particular values. If a system is unpredictable (high uncertainty), the entropy will be high and will be more uniform in its distribution. In our context, to assess the extent of clustering of the neural weights and eigenvalues, we computed the Shannon entropy of the neural weights and Spectral entropy of the eigenvalues. For our $N$ x $N$ recurrent weight matrices, we defined Shannon entropy of the weights, $W$, as:

$$H(W) = -\frac{1}{N}\sum_{i=1}^{N} p_{ij} log_2(p_{ij}), \text{ where } p_{ij} = \frac{w_{ij}}{\sum_{i=1}^{N} w_{ij}}$$

We quantified the Spectral entropy at the level of the eigenvalues of the weight matrix $W_\lambda$:

$$H(W_\lambda) = -\sum_{i=1}^{N} p_i log_2(p_i), \text{ where } p_i = \frac{|\lambda_i|}{\sum_{j=1}^{N} |\lambda_j|}$$

# Results
## Spatial and communication constraints enforce low entropy modularity in network weights

We find that both rate and spiking networks undergo marked changes in their topological features over the course of training (**Figure 2**). Prior findings demonstrate that rate seRNNs develop greater modularity relative to baseline models as they solve a one-step inference task[17]. Here, we find that this is also true for spiking seRNNs trained on the SHD task, with modularity also being greater in seRNNs relative to L1 networks (**Figure 2a**). This suggests that the topological structure of the network is driven by the constraints within the learning objective rather than as a feature of the underlying coding scheme or specific task.

While modularity demarcates the extent to which the network can be partitioned into sub-communities, it does not speak to the degree of uncertainty associated with the distribution of weights in the matrix. To assess this, we turned to the Shannon entropy of the weight matrix. Here, a uniform (or flat) probability distribution of weights would demonstrate a *high* Shannon entropy while a highly concentrated distribution, with some connections much stronger than others, would demonstrate a *low* Shannon entropy. We find that while there is a difference in the relative timing of the decrease in Shannon entropy over training, seRNNs form lower Shannon entropy networks compared to L1 networks irrespective of the coding scheme and task (**Figure 2b**). That is, the neural weights of seRNNs become increasingly concentrated within the weight matrix, leading to a lower Shannon entropy.

How does this correspond to the network's increased modularity? We find in **Figure 2c** that seRNNs exhibit clear negative linear relationship between the modularity and Shannon entropy, such that seRNNs with high modularity correspondingly exhibit low Shannon entropy and vice versa. This is not the case in L1 networks – modular L1 networks exhibit high Shannon entropy. Together, this suggests that while there may be many possible configurations of modular topologies (achievable via L1 regularisation during learning), seRNNs occupy a much smaller window of modular topologies, consistent with their constraints, as indicated by their low Shannon entropy. In what follows, we provide evidence suggesting this to be a general phenomenon of networks which learn under constraints.

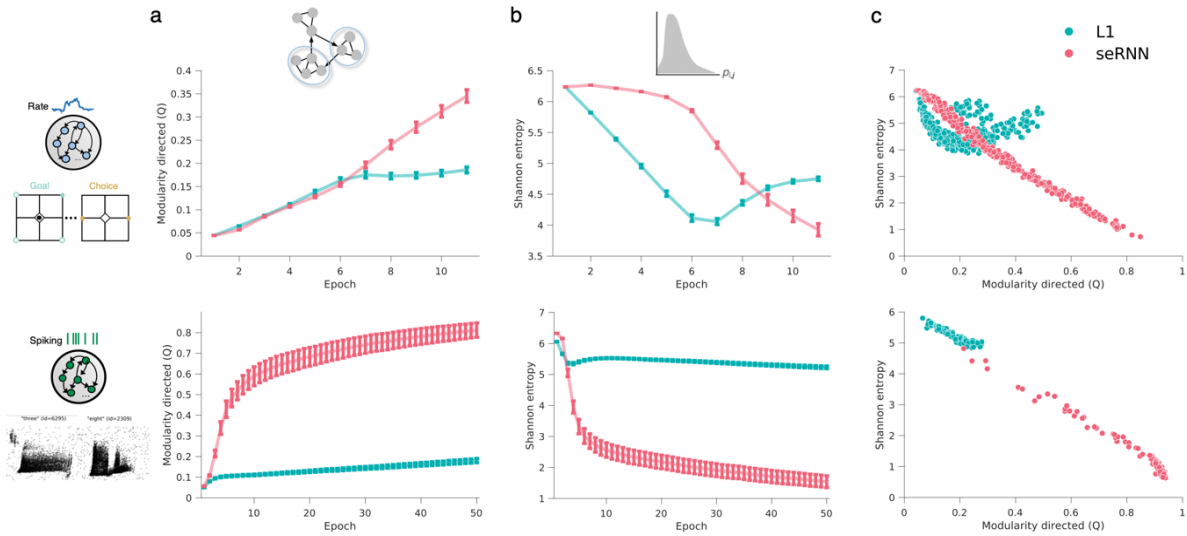

**Figure 2. Both rate and spiking seRNNs demonstrate low Shannon entropy modular networks.**
**a** Both seRNNs (pink) and L1 (green) networks increase in their directed modularity (Q) over the course of training in rate (top) and spiking (bottom) setups. For all plots, error bars correspond to two standard errors. **b** Correspondingly seRNNs achieve lower Shannon entropy by the end of training, with rate (top) networks achieving this later than spiking (bottom). **c** The relationship between directed modularity and Shannon entropy is linear for seRNNs but not for L1s, suggesting that seRNNs occupy a smaller number of low entropy modular configurations.

## Low entropy modularity is driven via an interpretable distance dependent connectivity and regular communicable topology

We have now seen that seRNNs develop low Shannon entropy modularity during training, indicative of weights becoming increasingly concentrated over the course of training. What next aimed to understand *why* by exploring what factors best explain the decrease in entropy.

Our initial findings are underscored by **Figure 3a**, which shows how relatively sparse networks, with decreased total neural weights, exhibit a much lower Shannon entropy compared to L1 networks. Further, in **Figure 3b** we show explicitly an example of a seRNN (**left, pink**) and L1 (**right, green**) network's probability distribution change over training, such that the seRNN's distribution become increasingly concentrated toward high frequency low probability transitions.

We find in **Figure 3c** that one core driving factor of this phenomenon is likely how space influences the concentration of weights. Specifically, in seRNNs, it is clear that already by mid-training (Epoch 5) probability values correlate directly with the length of that connection ($r = -0.427$, $p = 0.001$) and this is maintained into late training (Epoch 10) ($r = -0.284$, $p = 0.001$ respectively) (**Figure 3c, middle**). This does not occur in L1 networks as they are naturally unconstrained spatially (**Figure 3c, right**). We expect that one key reason for lower entropy modular configurations is that that space itself imposes a highly specified modular structure, promoted during the learning process, which is itself a single fixed configuration, which explains the decrease in Shannon entropy of the weights. We show in **Supplementary Figure 1** that, consistent with these findings, that networks constrained only in space achieve the lowest Shannon entropy compared to all other considered networks.

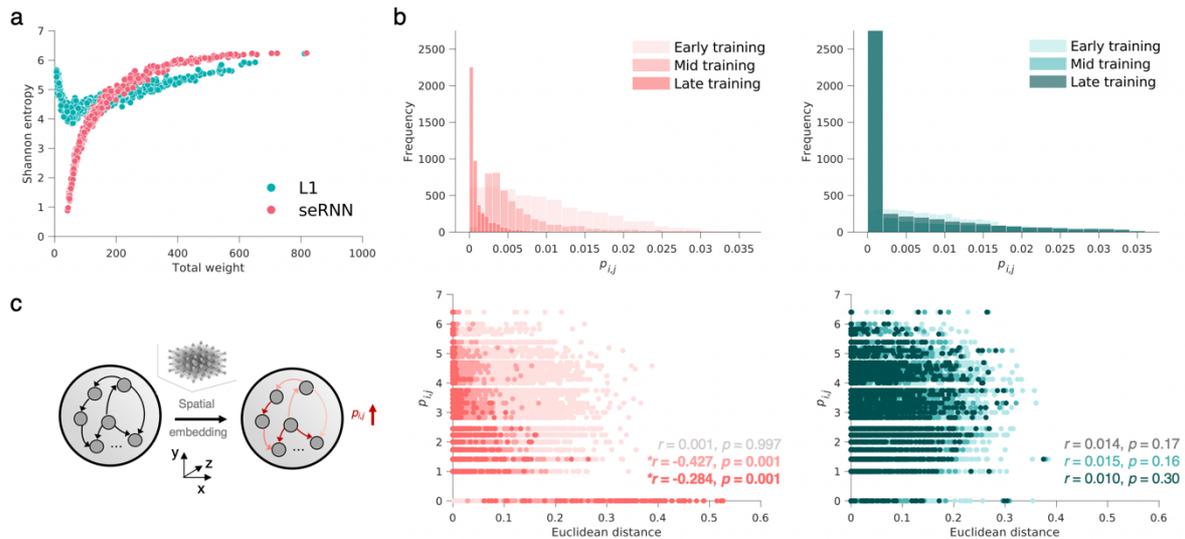

**Figure 3. Spatial embedding promotes interpretable distant-dependent decreases in Shannon entropy. a** The relationship between the total weight of the network and the Shannon entropy for seRNNs (pink) and L1 networks (green). **b** Network probability distributions ($p_{i,j}$), which derives from the weight matrix, of a representative seRNN (left, pink) and L1 network (right, green) over training (Epoch 1: Early training, Epoch 5: Mid training, Epoch 10: Late training). **c** A schematic illustration of how spatial embedding of seRNNs may explain concentrated probability distributions (left) which is demonstrated by scatter plots demonstrating how the neural weight $p_{i,j}$ is related intrinsically to the Euclidean distance between neurons (middle) but not L1 networks (right). Note in these plots, each point is a connection in the neural network.

While space is one key driver as explained above, seRNNs are also constrained in their communication via the network communicability. How does this manifest in terms of the network outcomes? We find that seRNNs, mirroring their weights, also generate lower Shannon entropy in their communicability matrices compared to L1 networks – suggesting that it is not only that the weight distribution of the network becomes more concentrated, but so too the communication within that network, as quantified by communicability (**Figure 4a**). This means that not only do weights become concentrated within more predictable connections based on space, but they also become more concentrated in the random walker movement through the network (as measured via network communicability). Overall, this suggests that seRNNs develop not only a low entropy modular spatial structure, but also topology of weights that generates low entropy communication pathways of random walkers (**Figure 4b**). Put simply, seRNNs develop a relatively small number of highly communicable connections, with an increasingly ordered topology, compared to L1 networks. The latter networks are unconstrained in terms of topology and hence remain relatively random and high entropy (**Figure 4c**).

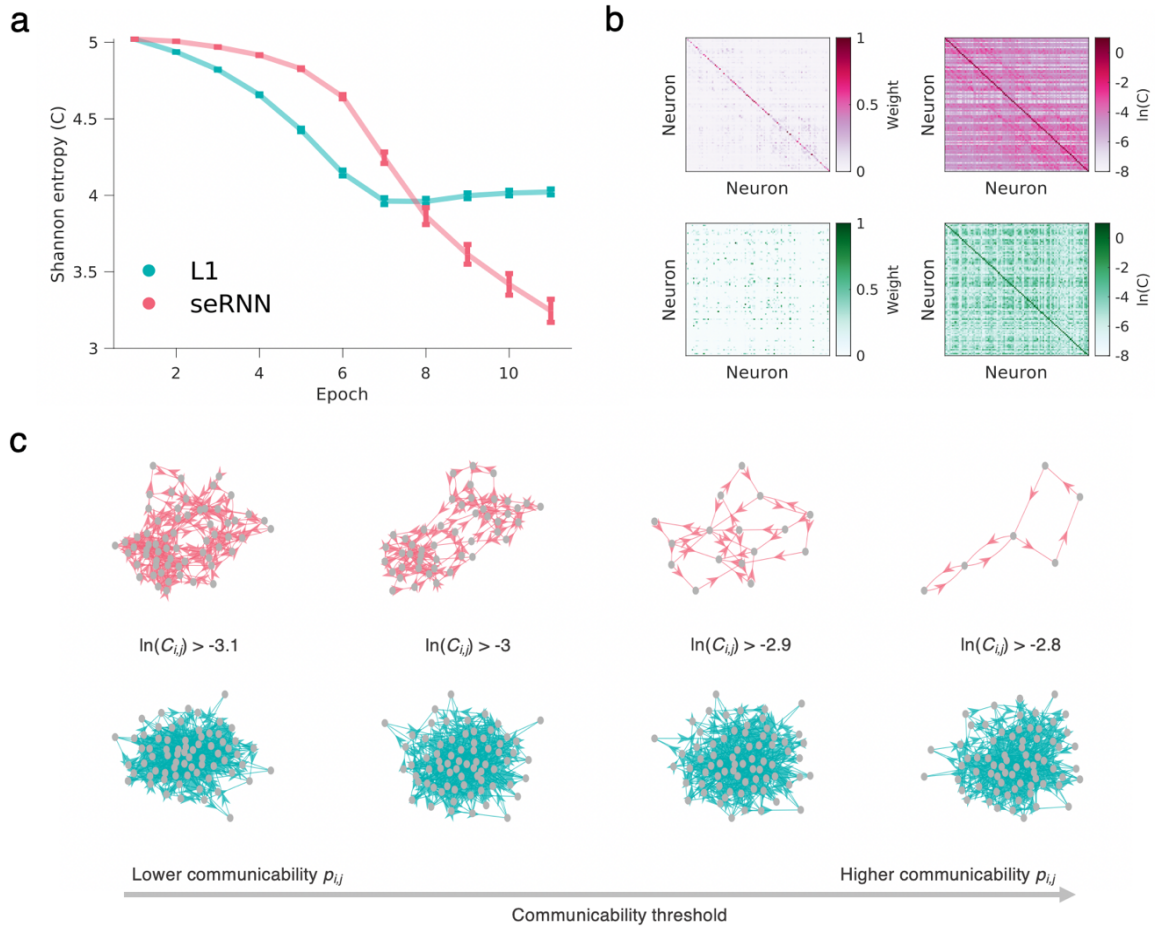

**Figure 4. Low entropy communicability generates regularity in communication pathways within seRNNs. a** Mirroring the decrease in Shannon entropy of the weights (as shown in Figure 2) seRNNs demonstrate decreased Shannon entropy in their communicability matrices, as demonstrated in example networks in **b** which shows the weight matrices of seRNNs (top, left) and L1 (bottom, left) in addition to their corresponding communicability matrices (right). **c** This demonstrates that for a set threshold of communicability, networks are more concentrated in their communicability for seRNNs meaning that their topological paths are more regular.

Together, our findings underscore a quite simple but important insight into the outcomes of constrained neural networks, as modelled by seRNNs: constraints operating on networks through learning bias network outcomes toward low entropy solutions. Importantly, these low entropy solutions may not have been clear without consideration of the constraints operating on the network during learning. Our findings show that any network may become modular, but only under particular constraints does this modularity become more specific and low Shannon entropy. In the case of seRNNs, weights exist on short connections in addition to connections that establish a relatively more regular, and hence predictable, topology for communication.

## Constraints directly influence the spectral dynamics of spatially embedded neural networks

So far, we have characterised the entropy of neural networks at the level of neural weights because this was the level at which we imposed constraints within the learning process. But what is the subsequent effect of these constraints on neural dynamics during task solving? This is an important question because it bridges directly the structural constraints placed on seRNNs with the resultant *functional capacity* of the resulting networks. To answer this, we turned to examining the eigenspectrum of our trained networks, which corresponds to the set of all eigenvalues of the networks, to elucidate the range of dynamic behaviour present in the networks over training. Crucially, we wished to examine if the constraints imposed on the neural networks, generating low entropy modularity in the weight space, also non-trivially changed the eigenspectrum.

The eigenspectrum can be used to understand numerous aspects of neural dynamics, including stability, intrinsic timescales and the prevalent directions of the dynamics[25,26,36,37]. We first investigated the eigenspectrum through two derivative measures across our considered seRNNs and L1 networks across both rate and spiking neural networks: the leading eigenvalue $\lambda_{max}$ and Spectral entropy, $H(W_\lambda)$. While the leading eigenvalue indicates the strength of dominant dynamics within the system, the Spectral entropy is a broader measure of the distribution of eigenvalues, indicative of the variability of dynamics present within the system (see *Methods* for more detail).

Our findings indicate marked differences in the eigenspectrum of seRNNs and L1 networks, irrespective of being rate or spiking or task. Specifically, in both rate and spiking setups, seRNNs generate a smaller $\lambda_{max}$ (**Figure 5a**) but higher Spectral entropy (**Figure 5b**) relative to L1 networks. These findings suggest that seRNNs demonstrate smaller dominant eigenvalues with, in general, more variability in the eigenvalues: indicating that constraints are not only affecting the network structures but are clearly doing so in a way that has downstream systematic consequences on the dynamics of the network.

To better understand what specifically is driving these spectral changes, we provide numerous eigenspectral visualisations showing how the eigenspectrum of seRNNs and L1 networks change depending on the level of regularisation strength imposed on the networks during learning (**Figure 5c**). As networks become increasingly constrained (going left to right), three observations become clear: seRNNs demonstrate increasingly (1) non-real eigenvalues (i.e., the eigenvalues collapse into the real axis only), (2) demonstrate increased variability in the eigenvalues along this real axis (corroborating an increased Spectral entropy relative to L1 networks) while (3) maintaining slightly smaller leading eigenvalues (corroborating a decreased $\lambda_{max}$ relative to L1 networks). The starkness of the differences of the eigenvalues between seRNNs and L1 networks suggests that under spatial and communication constraints, networks learn to solve the task under alternative dynamical configurations than would be possible otherwise (i.e., as in L1 networks). This is demonstrated through our first observation that the eigenvalues collapse their imaginary component of the eigenvalues, such that all the eigenvalues tend towards being all real (shown in **Figure 5c**). We propose that this is because, as spatial constraints are symmetrical (i.e., as the Euclidean distance between two neurons in three-dimensions is symmetrical), seRNNs correspondingly become increasingly symmetrical in their weight matrices compared to L1 networks, due to their preference to generate shorter connections. Importantly, this is known under spectral theorem: symmetrical matrices have all real eigenvalues[38].

In sum, our findings directly establish that constraints placed on networks during learning directly shape the eigenspectrum of the resultant networks, which is known to drive the dynamics of the underlying system. We suggest that this indicates that constraints do not necessarily have highly precise one-one effects on network outcomes. Instead, they limit the scope of possible configurations able to be achieved through learning on the network. We show this can span both structural and functional properties, despite our constraints here speaking only to structure. That is, the network exploits heterogeneity in its dynamics as a direct result of its structural constraints.

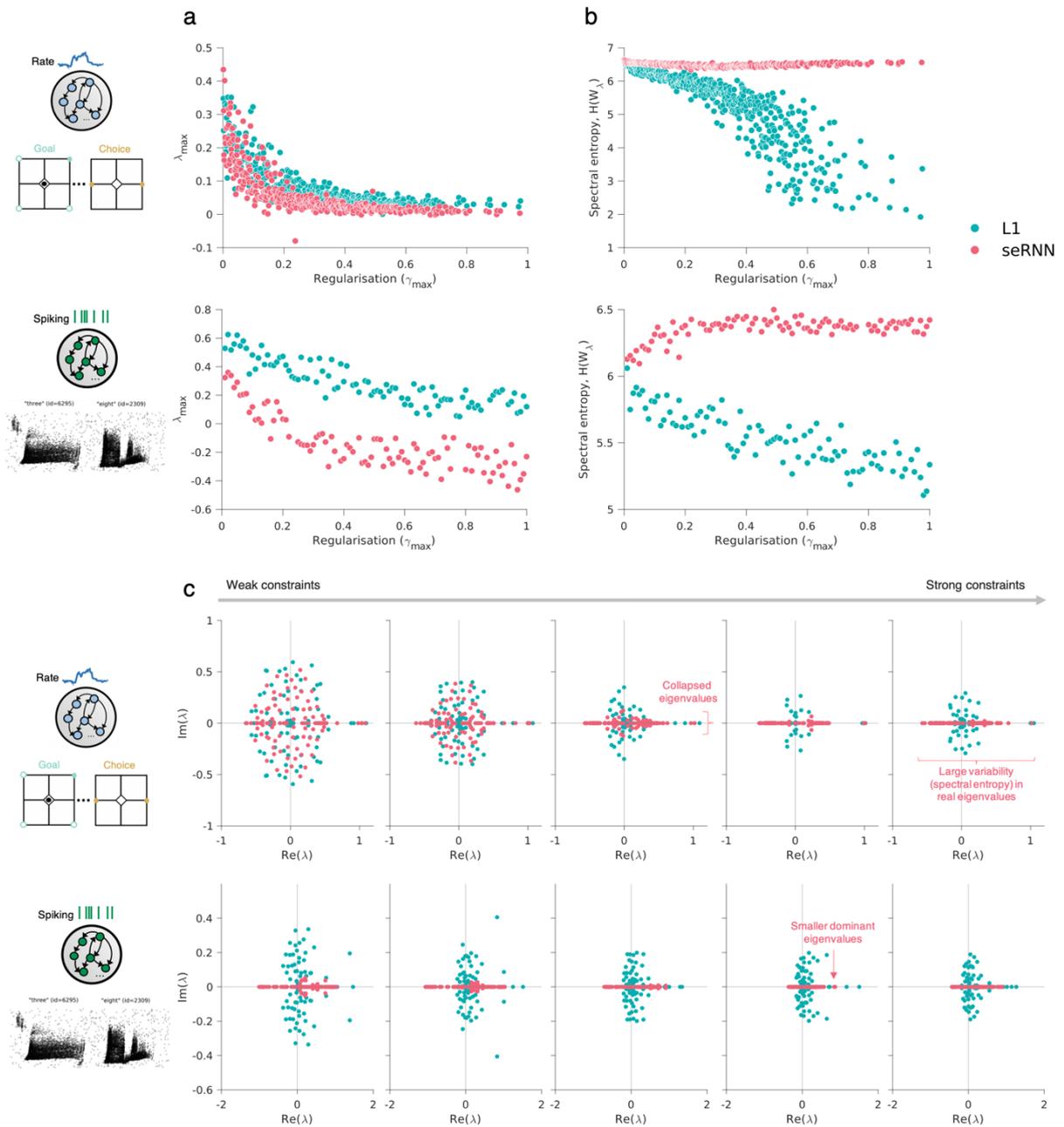

**Figure 5. Constraints influence the dynamics of trained neural networks, indicated through the network eigenspectrum. a** The relationship between the regularisation strength ($\gamma_{max}$) and the $\lambda_{max}$ for seRNNs (pink) and L1 networks (green) in trained networks in rate (top) and spiking (bottom) setups. Note the $\lambda_{max}$ is here plotted on a ln-scale. **b** The same plot but for the Spectral entropy. **c** The eigenspectrum for rate (top) and spiking (bottom) networks at 10% (left), 20% (middle left), 30% (middle), 40% (middle right) and 50% (right) of the maximum tested regularisation strength, where the eigenvalues (denoted, $\lambda$) of seRNNs and L1 are plotted. Note that eigenvalues here are complex, where eigenvalues have a real (x-axis) and imaginary component (y-axis). We highlight some key observations in seRNNs compared to L1 networks, including the imaginary component collapsing, large variation in the eigenvalues along the real axis and that the dominant eigenvalues in general remain smaller.

# Discussion

We have provided evidence that constraints imposed in rate and spiking seRNNs[17], across our modelled tasks through learning, lead to networks developing low entropy modular configurations. Importantly, this low entropy is explainable in terms of the structural constraints placed on the system: networks place their weights very specifically within relatively shorter connections (given spatial constraints) where it can also form regular, predictable connectivity (given local communication constraints). These changes in turn influence the dynamics of the system as indicated by the eigenspectrum[26] – drawing a direct line between the constraints experienced and dynamics available to the network[39]. Rather than only describing learned outcomes in artificial systems, the purpose of our work is understand concurrent network properties that emerge from constrained learning objectives[40], given that both task and connectivity must be understood together[41].

One example of new insight is our finding of the relationship between learned network spatial symmetries and the spectral theorem[38], where seRNN eigenvalues become increasingly real over learning. One compelling hypothesis stemming from this finding is that spatial constraints placed on biological networks bias their eigenspectrum toward having a greater spectral entropy (i.e., eigenvalues increasingly spread on the real axis), subsequently increasing the propensity toward more diverse timescales in the system dynamics[26,42]. However, we suggest this hints to a more fundamental point: networks exploit their available heterogeneity *when allowed to*. Under this view, the brain's deep heterogeneity occurs in tandem with its constraints[10,42,43]. This is in keeping with a growing body of work identifying the mathematical sequalae of various network constraints, such as sparsity[44], synaptic scaling[45], weight variances in learning[46] and Dale's law[37,44]. Put simply, complex dynamic phenomena can emerge from simple constraints that act upon a network as it learns.

There are a number of additional avenues for further work investigating constrained learning in seRNNs. For example, we investigated rate and spiking seRNNs across tasks, but these networks remain relatively small in size in terms of their learnable parameters. The scalability of this framework to encompass human geometry[1], delays[47], and multi-task setups[48] would all be highly fruitful avenues of exploration. Additionally, while we have pointed here toward promising directions in understanding the dynamical consequences of structural constraints, further work should look to more concretely formalise how varied constraints directly facilitate networks the ability to exploit their dynamical heterogeneity.

## Conclusions

Constraints instantiated via spatial embedding generate networks with low entropy modular configurations, with systematically altered eigenspectral properties, readily interpretable in light of the constraints placed on them. Our findings are consistent for rate and spiking networks across their respective tasks, speaking to a more general principle of exploited heterogeneity under constraints. Together, this work opens new directions for research seeking to establish how constraints limit the solution space in neural networks in computational neuroscience, where solutions to simultaneous structural and functional objectives must be accomplished in tandem.

## Acknowledgements

Both Duncan Astle and Danyal Akarca was supported by the Medical Research Council UK (MC-A0606-5PQ41), James S. McDonnell Foundation Opportunity Award for Understanding Human Cognition and the Templeton World Charitable Foundation (TWCF0159). Duncan Astle was supported by the Gnodde Goldman Sachs endowed Professorship in Neuroinformatics awarded to the University of Cambridge. Jascha Achterberg was supported by the Gates Foundation, Intel Labs, a project grant from the BBSRC (BB/X013340/1) and ERC-UKRI Frontier Research Guarantee Grant (EP/Y027841/1). Danyal Akarca was supported by the Imperial College Research Fellowship with support from Schmidt Sciences and a Nature Computes Better Opportunity Seed with the Advanced Research + Invention Agency (ARIA). For the purpose of open access, the authors have applied a CC BY public copyright licence to any Author Accepted Manuscript version arising from this submission.